# A Situation Calculus-based Approach To Model Ubiquitous Information Services[1]

Wenyu DONG, Ke XU, Mengxiang LIN

State Key Laboratory of Software Development Environment,

Computer Science & Engineering Department,

Beijing University of Aeronautics & Astronautics, Beijing, 100083, P. R. China

{wydong, kexu, mxlin}@nlsde.buaa.edu.cn

**Abstract:**    This paper presents an augmented situation calculus-based approach to model autonomous computing paradigm in ubiquitous information services. To make it practical for commercial development and easier to support autonomous paradigm imposed by ubiquitous information services, we made improvements based on Reiter's standard situation calculus. First we explore the inherent relationship between fluents and evolution: since not all fluents contribute to systems' evolution and some fluents can be derived from some others, we define those fluents that are sufficient and necessary to determine evolutional potential as decisive fluents, and then we prove that their successor states wrt to deterministic complex actions satisfy Markov property. Then, within the calculus framework we build, we introduce validity theory to model the autonomous services with application-specific validity requirements, including: validity fluents to axiomatize validity requirements, heuristic multiple alternative service choices ranging from complete acceptance, partial acceptance, to complete rejection, and validity-ensured policy to comprise such alternative service choices into organic, autonomously-computable services. Our approach is demonstrated by a ubiquitous calendaring service, *ACS*, throughout the paper.

## 1    Ubiquitous Information Service And Modeling Challenge

### Ubiquitous Information Service

With booming telecommunication and embedded computing technologies, *ubiquitous application* [1] is believed to come in prevalence in one or two decades [2][3]. By then, traditional Web-based information services will be augmented into ubiquitous ones, i.e. they are accessible from desktop systems as well as from embedded computing terminals, via wired as well as via wireless networks, by sedentary as well as by mobile users whenever and wherever they wish, as depicted in figure 1-1.

WAP (Wireless Application Protocol) services are prototypes of ubiquitous information services. Since 2000, many WAP services have been set up, but unfortunately almost none of them have achieved revolutionary success as expected.

The reason lies in the collision between service scenario and accessing context. WAP endows Web-based information services with mobile accessing, but copies almost the same service scenario from traditional Web-base information services. The service scenario is driven by, and heavily relies on, users' I/O operations, basically a series of read/write-intensive actions. Unfortunately, for nowadays WAP services, esp. those using mobile phones as accessing terminals, there exist several seemingly un-surmountable difficulties:

- *Physical limitation:*    Because of in-born portable requirements, mobile phones are allowed only limited weight, size and power consumption, and thus unable to support as powerful human-machine interaction facilities as desktop machines do. For example, desktop systems support powerful GUI interaction with 103/104-key keyboards, 15/17-inch monitors and mice, while mobile phones usually have only primitive dial keyboards and much smaller monitors.

- *Physiological limitation:*    Human beings are not physiologically suitable to perform reading- and/or writing-intensive activities in movement [4]; otherwise mobile humans might catch in-convenience, if not danger, for example one who is driving.

[1]  This research is supported by China's National Basic Research Program (973 Program) grant No.G1999032701 and the second author is supported by FANEDD grant No. 200241.

An intuitive solution out of this dilemma is to develop highly-autonomous service scenario: For real-time interactions, try the best to augment service logic to be autonomously-computable, i.e. computable without users' interaction; and/or try the best to postpone the computing to a later time, waiting for the user to become convenient to use traditional desktop systems, for example when he returns to his office and log on his desktop PC.

Fortunately this can be done for many information services. Consider calendaring service, a typical network PIM (Personal Information Management) application: it can be augmented to a ubiquitous service called *Active Calendar Service*[2], or *ACS* in short hereafter. The user's agenda schedule is stored in a network server, which will alert the user of the on-going events. ACS runs on a ubiquitous network, as depicted in figure 1-1 wherein ACS runs on server A.

Consider the exemplar events and the service scenario in figure 1-2. Suppose that during the scenario in figure 1-2, the user is in movement, thus he is in-convenient to use perform real-time, human-machine interactions and has to rely on the autonomy of ACS service.

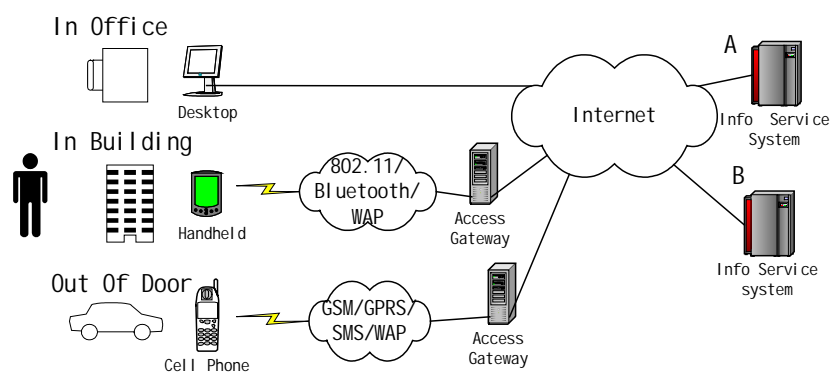

Fig.1-1    ubiquitous information services

By time $t_0$ the user has been arranged a schedule {*ev001*, *ev002*}. At time $t_1$ a request comes, say from B in figure 1-1, to add event *ev003*, which conflicts with *ev002* in time duration but has higher importance level ('*high*' vs. '*normal*'). So it's better to have the user to make the final decision (acceptance or rejection). Since the user is now in movement and is in-convenient to do so, ACS exploits a wiser, autonomous policy: it temporarily stores *ev003*, waiting for the user to make the final decision at a later, convenient time when, for example, he becomes convenient to use desktop systems.

---

[2]  Shortened as *ACS* hereafter.

Thus we can see that ACS has prepared multiple choices to serve the request of adding *ev003*, and at this time the request is partially accepted, neither completely accepted nor completely rejected. Similarly, at time $t_2$ *ev004* is temporarily stored since it conflicts with, and has higher importance level than, *ev001*.

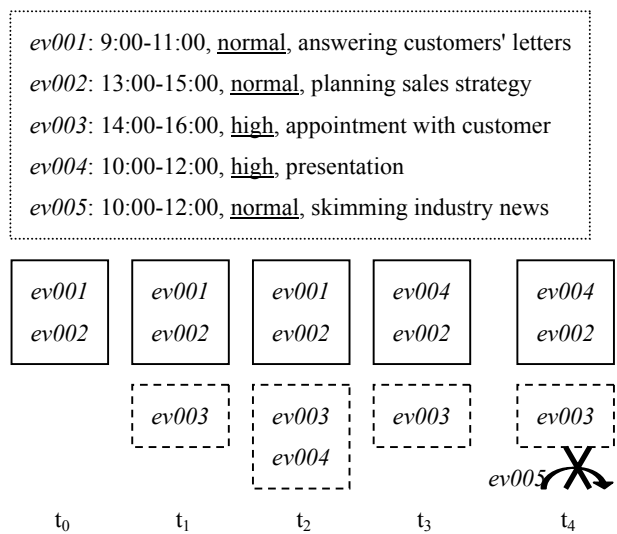

Fig.1-2        agenda arrangement scenario

We call the above policy as *Prioritized Multi-Purpose* policy[3]: multiple choices are available to serve a request: completely satisfied, partly satisfied, completely rejected and something else, and all alternative approaches are selected to execute on priority basis. Usually the approach with the highest priority results in complete satisfaction; the approaches with intermediate priority result in half satisfaction and half rejection (for example at time $t_1$ in the scenario in figure 1-2); and the approach with lowest priority results in completely rejection (for example at time $t_5$ in the scenario in figure 1-2). Some application-dependent criteria must be obeyed when selecting the service choices. For example in ACS the criteria might be: there is no time duration overlapping between completely accepted events, and an event is allowed to temporarily stored only if it overlaps in time-duration with at least one completely accepted event but has higher importance level.

At time $t_3$ for some reasons *ev001* is cancelled, for example someone else takes over to answer the letters. Since no event conflicts with *ev004*, it should be automatically re-arranged as a completely accepted on-going event. The re-arrangement of *ev004* can be done *autonomously*, free from users' intervention.

From the above example it can be seen that addition of *ev003* and *ev004*, deletion of *ev001* and re-arrangement of *ev004* are handled independent of the user's intervention, and that ACS works far beyond than just following the user's orders. The service scenario is distinct from that in traditional Web-based information services.

## Modeling Challenge

It is theoretically possible to build such models as that of ACS with simple, common-used FSM (Finite State Machine[4]), but FSM often leads to programs that are hard to write, debug, and maintain [5].

---

[3] Shortened as *PMP* hereafter.
[4] Shortened as *FSM* hereafter.

FSM is a 'flat' modeling formalism. Primarily it has two sorts: states and transitions, and usually they are structured un-hierarchically: all states are presented in the same layer; transitions are free to connect any pair of states, and thus converge, concatenate or cross over one another in a un-sorted way. Thus an FSM often appears like a mass of spaghetti in the end.

Besides, FSM is a case-specific solution. To develop a new application, usually the FSM will undergo completely re-designing and re-constructing. When requirement changes, the states and transitions are coupled so tightly that slight change often leads to substantial amount of maintenance work.

Thirdly, FSM requires enumeration of all possible states in advance. Sometimes such enumeration is impossible, or will cause complexity problem [5]. Consider ACS scenario in figure 1-2, a machine state should be represented by combination of all stored events. During design phase, it is impossible to enumerate all possible events that might appear during run-time phase: it is always possible that there comes an event out of designers' anticipation. Even if enumeration is possible, changes in state composition would cause an exponential increase in the number of machine states [5].

We appeal to situation calculus to deal with the modeling challenge. Situation calculus [6] has long been the formalism in modeling dynamic world, covering robotics, database updates, agent programming [7][8][9][10][11], etc. In such applications computer systems are often required to handle complex interaction and autonomous computing, just as required by ubiquitous information services. Proven success makes situation calculus promising to model ubiquitous information services.

The purpose of this paper is to explore the usefulness of situation calculus in modeling commercialization-ready ubiquitous information services. Situation calculus is used differently in these applications from those traditional AI problems such as reasoning, planning, etc: situation calculus is going to be augmented, integrated in and combined into commercial development methodology.

In the next section, we start with reviewing standard situation calculus. Then we explore in detail the role of fluents and situations so as to revise situation calculus to become eligible for development methodology. Then we propose a heuristic validity theory to model autonomous computing paradigm for ubiquitous information services. The exemplar ACS[5] service is examined throughout this paper, and complete calculus description of the model and the service scenario in figure 1-2 are demonstrated in the end.

## 2   Informal Review for Situation Calculus

### 2.1   Informal Ontology

A situation calculus [7][8] is a many-sorted first-order language (with some second-order features) specifically designed for representing dynamically changing worlds. All changes to the world are the result of a sort *ACTION*. A sort *SITUATION* represents the complete states of the world at an instant of time, and is simply a sequence of actions in Reiter's formalism. A sort

---

[5]   See page 2 for more description of ACS (Active Calendaring Service).

*OBJECT* catches everything else depending on the application domain. A constant $S_0$ is used to denote the initial situation. A distinguished binary function symbol $do(a,s)$ denotes the situation resulting from performing action $a$ in situation $s$. A binary predicate symbol $<$: $s_1<s_2$ means $s_2$ can be obtained from $s_1$ by performing a certain sequence of actions. There are finitely many situation-independent predicates with arity $(ACTION \bigcup OBJECT)^n$ ($n{\geq}0$), finitely many situation-independent functions with arity $(ACTION \bigcup OBJECT)^n \rightarrow OBJECT$ ($n{\geq}0$), finitely many *relational fluent*s which are presdicates with truth values varying from situation to situation and with arity $(ACTION \bigcup OBJECT)^n \times SITUATION$ ($n{\geq}0$), and finitely many *functional fluents*[6] which are function symbols with values varying from situation to situation and with arity $(ACTION \bigcup OBJECT)^n \times SITUATION \rightarrow OBJECT$ ($n{\geq}0$).

## 2.2   Actions and Effects

Every action is specified by a *Pre-condition Axiom* and *Successor State Axiom*. The former states the prerequisite under which it can be performed:

$$Poss(A(x_1,\cdots,x_n),s) \equiv \Pi_A(x_1,\cdots,x_n,s) ,  \tag{2-1}$$

where $\Pi_A$ is a formula uniform in $s$[7] and whose free variables are among $x_1,..., x_n$ and $s$.

A *Successor State Axiom* states the how the action affects the world, namely changing in fluents:

$$F(x_1,\cdots,x_n,do(a,s)) \equiv \Phi_F(x_1,\cdots,x_n,a,s) ,  \tag{2-2}$$

$$f(x_1,\cdots,x_n,do(a,s)) = y \equiv \phi_f(x_1,\cdots,x_n,y,a,s) ,  \tag{2-3}$$

where $F$ is a ($n$+1)-ary relational fluent, and $\Phi_F$ is a formula uniform in s and whose free variables are among $x_1,...,x_n$, $a$ and $s$; $f$ is a ($n$+1)-ary functional fluent, and $\varphi_f$ is a formula uniform in s and whose free variables are among $x_1,...,x_n$, $y$, $a$ and $s$.

## 2.3   Complex Actions

To express and reason with complex actions, Reiter introduced complex action theory and implemented a high-level programming language GOLOG [8][12], which supports:

(1)   Primitive actions
(2)   *Test* action: $\varphi$?
(3)   *Sequence*: $[\delta_1; \delta_2]$.
(4)   *Non-deterministic choice of actions*: $[\delta_1|\delta_2]$.
(5)   *Non-deterministic choice of action arguments*: $[\pi x. \delta(x)]$.
(6)   *Non-deterministic iteration*: $\delta^*$.
(7)   *Conditional*: **if** $\varphi$ **then** $\delta_1$ **else** $\delta_2$ **endif**.

---

[6]   For clarity purpose, here we assume fluents would not produce an action as its result.
[7]   see [6] for more about "uniform in s".

(8)  *Loop*: **while** $\varphi$ **do** $\delta$ **endwhile**.
(9)  *Procedure definition*: **proc** $\beta(x)$ $\delta$.

## 2.4    Discussions

### 2.4.1    Merit

Reiter's situation calculus has received honorable recognition and become one of the de-facto modeling formalisms for dynamic worlds.

Physical worlds are regarded as infinitely many successive transitions (i.e. actions) between infinitely many snapshots (situations). Situations are formally defined by Reiter as sequences of actions executed. Transitions are implied by relevant prerequisites and changes in situation-dependent properties (i.e. fluents). Reiter proposed a straightforward and working solution to frame problem, and complex action theory to develop high-level program.

The underlying philosophy of situation calculus is similar to common-used FSM. Situation calculus surpasses FSM in that it eliminates the in-advance enumeration of states prior to running time. Situations are dynamically generated during running time and can have infinitely many instances, making situation calculus eligible to model open, dynamic world.

Another merit is that its calculus formalism conveys richer description capability. Pre-condition axioms are used to ensure correct execution of actions, successor state axioms are used to represent effects of execution, and complex action theory are used to express sequential, conditional and looping execution series. All these are built on formal calculus.

### 2.4.2    Difficulty

However standard situation calculus is not suited enough to serve as modeling formalism for commercial development of ubiquitous information services, since the formal calculus is not compatible with, and is hard to be mapped into, practical development methodology.

Maybe the hardest under-development of standard situation calculus is about evolution (or progression). To avoid Reiter's action-sequence-based situations might ever increase in length, the calculus system must periodically 'forget' the past. Fluents are chosen as media via which to associate a later situation with an older one: Lin believes that two models are equivalent if they agree on all fluents [10][11]; and Thielscher composes state update axioms as blind combination of every fluent that might have positive and negative transition [13][14].

However we believe it is too coarse to treat all fluents as equally important. First, some situations can be believed equivalent to a certain extent though they may not agree on some trivial fluents. Supposing ACS has a fluent *time(s)* that tells the time of universe, and ACS makes no use of it except as time stamp for logging purpose. Considering time $t_3$ and $t_4$ in figure 1-2, although they do not agree on *time(s)*, the two relevant situations are the same from functional point of view. Second, usually fluents are not independent of one another. For example in blocks world [7][15], we have $clear(x,s) \equiv \forall y.\neg on(y,x,s)$. Then at each evolution step, it is not necessary to compute *clear*() for pure progression purpose as long as *on*() is always done. Cutting-down in the number of fluents that need computing during evolution will relieve much complexity burden. Finally but not least, as far as we know, traditional fluent- or state-based approaches are often built on set of relational fluents, while functional fluents are completely omitted [10][11][13][14]. Although this may seem trivial in theoretical point of view and one might build explicit mapping functions to bridge them, this may cause much extra work in practical development if standard situation calculus is used as direct modeling reference.

Another in-capability is entailed by PMP[8] policy. As shown in ACS example, PMP is often required in ubiquitous information service. Basically PMP needs to describe application-dependent criteria, multiple service choices and the decision strategy by which to choose one of the choices to execute. Standard situation calculus lacks relevant mechanism to describe, to customize and to reason PMP policy, or at least is inconvenient to do so. In most proven cases [7][8][9][10][11] of situation calculus, if the raw request cannot be completely satisfied, it will be just completely rejected at all.

The following section is devoted to explore the mapping problem. Then we introduce Validity Theory to model PMP policy.

# 3   Fluent and Situation

In standard situation calculus, fluents are literally treated dummy storage to record execution result of actions. Later in Lin [10][11] and Thielscher [13][14] fluents are used to deduce evolution or progression. In fact more can be done with fluents since they are inherent associated with situation and evolution.

McCarthy [16] regards situation as the complete set of states of the universe at an instant of time. For any given application, it is never possible, nor necessary, to do so, since only finitely many properties are involved and they are sufficient to deduce relevant properties in current situation and in later ones. The question is: what properties should be selected? How do they function in situation calculus?

From pragmatic point of view, a situation marks a timing point during the system's (theoretically endless) evolution procedure, thus, if only future is considered[9], a situation can be specified only by its evolution potential from that time on. Since the procedure composed of a sequence of actions executed, and since the pre-conditions in every intermediate situation determine the execution possibility at the next step, pre-conditions become the entrance to study evolution procedure.

Moreover, as shown in the *blocks* world [7][15], pre-conditions can often be derived from fluents, e.g. $Poss(pickup(x),s) \equiv nothing\_in\_hand(s) \wedge heavy(x,s)$. That is, fluents may have underlying but severe impact on evolution, and detailed exploration is needed here.

## 3.1   Preliminaries

We start with introduction of some auxiliary symbols and conceptions.

---

[8]  See page 3 for more about PMP, or Prioritized Multi-Purpose policy.
[9]  It is reasonable to omit history since in most cases people care more about behavior in future evolution than the history record.

### 3.1.1    TESTACTION

Assume in a given situation calculus system, there can be only finitely many test actions: $\varphi_1?,\ldots,\varphi_n?$ (n≥0). We introduce a new sort: **TESTACTION** = $\{\varphi_1,\ldots,\varphi_n\}$ (n≥0)[10]

### 3.1.2    Derivation Symbol

Suppose $F$, $G_1$, …, $G_m$ are all functions, then:

Let $F \leftarrow_F \{G_1,\cdots,G_m\}$ denote that $F$ can be derived from $\{G_1,\ldots,G_m\}$ (m≥0), i.e. there exist a function $func_F$ such that $F = func_F(G_1,\cdots,G_m)$.

Let $F \leftarrow_F^{nec} \{G_1,\cdots,G_m\}$ denote that each of $\{G_1,\ldots,G_m\}$ is necessary to derive $F$, i.e.

$$F \leftarrow_F \{G_1,\cdots,G_m\} \text{ and } \forall i,1\leq i\leq m.\neg[F \leftarrow_F \{G_1,\cdots,G_{i-1},G_{i+1},\cdots,G_m\}].$$

### 3.1.3    Seed Set and Atomic Fluent

We define:

**Def 3.1    Seed Set**  A *Seed Set* of a fluent $F$ is a fluent set: $\{G_1,\ldots,G_m\}$, $G_i\neq F$, (1≤i≤m), such that $F \leftarrow_F^{nec} \{G_1,\cdots,G_m\}$.                                      ∎

Usually fluents are not independent from one another. Some of them can be derived from some others. For example, in the blocks world, fluent $clear(x,s)$ can be derived from $on(y,x,s)$: $clear(x,s) \equiv \forall y.\neg on(y,x,s)$. That is, $clear(x,s)$ has a seed set $\{on(y,x,s)\}$.

Obviously a fluent might have multiple seed sets, and a fluent might be in multiple fluents' seed sets. For example, in a fluent set $FS=\{F_1, F_2, G_1\}$, derivation relations are $\{G_1\leftarrow_{G11}\{F_1\}$, $G_1\leftarrow_{G12}\{F_2\}\}$. $G_1$'s seed set might be either $\{F_1\}$ or $\{F_2\}$.

**Def 3.2    Atomic Fluent**      A fluent is called an *Atomic Fluent* if it has no seed set but an empty one.                                                                                        ∎

*Atomic fluent*s are those that cannot be derived from other fluents. Intuitively, they can be used to encapsulate raw information such as jumbo data storage, physical sensors or other network services out of this system, *non-atomic fluent*s may be derived from atomic ones and/or other non-atomic ones by logic computing. For example, $clear(x,s)$ is non-atomic, and $on(x,y,s)$ will be atomic if it cannot be derived from other fluents.

### 3.1.4    Generalized Fluents

For clarity purpose, we generalize fluent's notation to cover pre-conditions, relational fluents and functional fluents.

---

[10]  This is reasonable since in most practical applications, their complex actions, including test actions, can be enumerated in requirement analysis phase or design phase, prior to running time.

3.1.4.1    Precondition and Fluent

Consider formula 2-1. Literally $\Pi_A$, or $Poss(a,s)$, is a predicate with arity $(ACTION \bigcup OBJECT)^n \times SITUATION$ ($n{\geq}0$), just the same as that of relational fluents. Actually, pre-conditions can often been derived from fluents. For example, in the *blocks* world [7][16] there is a pre-condition:

$Poss(pickup(x),s) \equiv nothing\_in\_hand(s) \wedge heavy(x,s)$ .

The precondition of *pickup(x)* can be derived from *nothing_in_hand(s)* and *heavy(x,s)*.

We believe this is the case for every action in every situation calculus system. Thus, in this paper we treat pre-conditions, $\Pi_A$ or Poss($A(x),s$), as relational fluents, and we also apply the conceptions of seed set and atomic fluent onto pre-conditions.

3.1.4.2    Relational And Functional Fluent

Suppose the sort OBJECT contains Boolean type, i.e. $\{true, false\} \subseteq OBJECT$ . Then a relational fluent can be considered as a functional one with $\{true, false\}$ as its value domain. In the rest of this paper, unless explicitly mentioned, we do not distinguish relational fluents from functional ones. All fluents will be treated as functional ones, denoted as $F(x_1,\ldots, x_n, s)$.

Similarly, consider successor state formula 2-3. Suppose there is a function $\Phi_f$ such that $\Phi_f(x_1,\cdots,x_n,a,s) = y$ . Then the successor state for fluent $f$ can be re-written as $f(x_1,\cdots,x_n,do(a,s)) = \Phi_f(x_1,\cdots,x_n,a,s)$ . This is very similar to formula 2-2. In general, for any fluent, no matter relational or functional, its successor state axiom can be written as:

$$F(x_1,\cdots,x_n,do(a,s)) = \Phi_F(x_1,\cdots,x_n,a,s) , \qquad\qquad (3\text{-}1)$$

where $F$ is a (relational or functional) fluent; and $\Phi_F$ is a formula uniform in $s$ with the same value domain as that of $F$ and with free variables are among $x_1,\ldots,x_n$, $a$ and $s$.

### 3.1.5    Deterministic Complex Action

The complex actions defined in GOLOG [8][12] imply some indeterminacy. To facilitate later discussion, we remove the indeterminacy.

First we (informally) introduce:

**Deterministic Choice of Actions** [$\delta_1 >> \delta_2$]: $\delta_1 >> \delta_2$ means $\delta_2$ is executed only if $\delta_1$ cannot be.

**Deterministic Choice of Action Arguments** [$\rho\pi(x).\delta(x)$]: In situation $s$, suppose all $x$ that satisfies $Poss(\delta(x)) \equiv$ true can be ordered according to certain kind of priority, say $x_1, x_2, \ldots, x_n, \ldots$, then $\rho\pi(x).\delta(x)$ means $\delta(x_1) >> \delta(x_2) >> \ldots >> \delta(x_n) >> \ldots$.

**Deterministic Iteration** [$\delta^\infty$]: $\delta^\infty$ means $\delta$ is executed over and over and execution will not cease until $Poss(\delta)$ becomes false.

Then we define:

**Def 3.4    Deterministic Complex Action**    A *Deterministic Complex Action* can be iteratively

composed from:

(1)   Primitive action,

(2)   *Test* action: $[\varphi?]$,

(3)   Sequence: $[\delta_1;\delta_2]$,

(4)   *Deterministic choice of actions*: $[\delta_1 >> \delta_2]$,

(5)   *Deterministic choice of action arguments*: $[\rho\pi(x).\delta(x)]$, or

(6)   *Deterministic iteration*: $\delta^\infty$.

Where $\delta$, $\delta_1$ and $\delta_2$ are deterministic complex actions.                                    ∎

Conditionals and while-loops can be defined in terms of the above constructs:

**if** $\varphi$ **then** $\delta_1$ **else** $\delta_2$ **endif** $= [\varphi?;\delta_1] >> [(\text{not } \varphi?);\delta_2]$, and

**while** $\varphi$ **do** $\delta$ **endwhile** $= [\varphi?;\delta_1]^\infty$.

Elimination of in-determinacy does not handicap deterministic complex actions to describe commercial information services. More likely the elimination will facilitate or simplify the description, since most commercial development platforms and utilities do not support in-deterministic reasoning as Prolog or GOLOG does.

### 3.1.6    Closeness of Successor State

In standard situation calculus, little restriction is specified for successor state axioms (except that $\Phi_F$ in formula 3-1 should be uniform in $s$ with free variables among $a$, $a$'s parameters and $s$). In practical applications, there might be quite a number of fluents whose successor states can be derived from the action's parameters and some fluents' value in the current situation. For example, let a fluent *table_emp*$(s)$ denote the database table containing all employee records, we have:

$$Poss(add(emp), s) \supset table\_emp(do(add(emp), s)) = table\_emp(s) \bigcup \{emp\}.$$

In blocks world, let *on_pairs*$(s)$ denote $\{(x,y)|x \text{ is on } y \text{ adjacently}\}$. Then:

$$Poss(move(u,v,w), s) \supset$$
$$on\_pairs(do(move(u,v,w), s)) = on\_pairs(s) - \{(u,v)\} \bigcup \{(u,w)\}.$$

Similarly in ACS, let *eventlist*$(s)$ denote all events in the schedule, we have:

$$Poss(add(ev), s) \supset eventlist(do(add(ev), s)) = eventlist(s) \bigcup \{ev\}.$$

We define:

**Def 3.3   Close Successor State** and **Incremental Successor State**   Say a fluent *F's successor state wrt action a is closed wrt to a fluent set* $C = \{G_1, \ldots, G_n\}$  $G_i \in FLUENT$  $(n \geq 0)$, if there exists a function $\Psi_{F,a}$ such that:

$$Poss(a(x), s) \supset F(do(a(x), s)) = \Psi_{F,a}(x, G_1(s), \cdots, G_n(s)).  \tag{3-2}$$

Specially, if $C = \{F\}$, i.e.

$$Poss(a(x), s) \supset F(do(a(x), s)) = \Psi_{F,a}(x, F(s)).  \tag{3-3}$$

Formula 3-3 is called an *incremental successor state* axiom, and $F$ is called an *incremental fluetn wrt action a*.                                    ∎

Usually the incremental fluent's successor state is obtained from partial modification on the existing value. The modification is close with the action executed. This is common in database operation, as shown in the examples above. Incremental fluents are easy to compute and common in practical applications, thus convey much directive significance.

## 3.2    Situation Equivalence

In this section, situation is formally studied from 'pragmatic' point of view. Since situations are basically time points during evolution, esp. for future, we define:

**Def 3.5    Evolutional Equivalence for Situations**    Say situation $s_1$ and $s_2$ are **evolutionally equivalent** if for any deterministic complex action $\delta$ of arbitrary length (including zero):

(1)  $\forall a \in ACTION.Poss(a,do(\delta,s_1)) \equiv Poss(a,do(\delta,s_2))$ , and

(2)  $\forall \phi \in TESTACTION.\phi[do(\delta,s_1)] \equiv \phi[do(\delta,s_2)]$ .                    ∎

Roughly situations of evolutional equivalence have the same evolution potential in arbitrary far future. Inductively, after executing the same task (the same decisive complex action $\delta$), they still have the same possibility to execute any action at the next step.

Evolutional equivalence does not necessarily ensure that the system will exhibit the same functions to outside world. For example, some fluents that do not affect evolution might have different values in the starting situations and the difference is retained during the evolution procedure so far. Thus we define:

**Def 3.6    Functional Equivalence for Situations**    Say situation $s_1$ and $s_2$ are **functionally equivalent** if for any deterministic complex action $\delta$ of arbitrary length (including zero):

(1)  $s_1$ and $s_2$ are evolutionally equivalent, and

(2)  $\forall F \in FLUENT.F(x,do(\delta,s_1)) \equiv F(x,do(\delta,s_2))$ .                    ∎

We have a straightforward proposition:

**Proposition 3.1**    Suppose every action's precondition can be derived from fluents, i.e.

$\forall a \in ACTION, \exists F_1 \cdots F_n \in FLUENT.Poss(a,s) \leftarrow_{Poss(a)} \{F_1, \cdots, F_n\}$ ,

$s_1$ and $s_2$ are evolutionally equivalent (and hence functionally equivalent) if, for any deterministic complex action $\delta$ of arbitrary length (including zero), do($\delta,s_1$) and do($\delta,s_2$) agree on:

(1)  every fluent in FLUENT, and

(2)  every test action in TESTACTION.                    ∎

**Proof**:    Straightforward.                    ∎

## 3.3    Decisive Fluent and Non-decisive Fluent

Intuitively, those fluents that affect pre-conditions or test actions may have more impact on the system's evolution; moreover, fluents are not independent one another, some of them may be derived from some others. In this section, we will choose the most *important* fluents by combining these two criteria.

**Def 3.7    Precondition Fluent Base**    The *Precondition Fluent Base*, denoted as $\overrightarrow{PFB}$ , is a subset of *FLUENT* such that:

1) $\forall F \in PFB.[\exists a \in ACTION, \exists \vec{G} \subseteq PFB.[Poss(a) \leftarrow^{nec}_{Poss(a)} (\vec{G} \cup \{F\})]]$ [11], and

$\lor [\exists \delta \in TESTACTION, \exists \vec{G} \subseteq PFB.[\delta[s] \leftarrow^{nec}_{\delta} (\vec{G} \cup \{F\})]]$

2) $\forall F \in (FLUENT - PFB).[\forall a \in ACTION, \forall \vec{G} \subseteq PFB.\neg[Poss(a) \leftarrow^{nec}_{Poss(a)} (\vec{G} \cup \{F\})]]$

$\land [\forall \delta \in TESTACTION, \forall \vec{G} \subseteq PFB.\neg[\delta[s] \leftarrow^{nec}_{\delta} (\vec{G} \cup \{F\})]]$ ∎

**Def 3.8** **Decisive Fluent Set**, **Decisive Fluent** and **Non-decisive Fluent** For a given situation calculus system, its *Decisive Fluent Set*, denoted as $\vec{DF}$, is a subset of $\vec{PFB}$ such that:

(1) $\forall G \in \vec{NF}, \exists \{F_1, \cdots, F_n\} \subseteq \vec{DF}, n > 0.G \leftarrow_G \{F_1, \cdots, F_n\}$ (where $\vec{NF} \overset{def}{=} \vec{PFB} - \vec{DF}$), and

(2) $\forall F \in \vec{DF}, \exists G \in \vec{NF}.\neg[G \leftarrow_G (\vec{DF} - \{F\})]$.

A fluent $F$ is called a *Decisive Fluent* if $F \in \vec{DF}$; and is called a *Non-decisive* one if $F \in \vec{NF}$. ∎

The value tuple of $\vec{DF}$ in situation $s$ is denoted as $\vec{DF}[s] \overset{def}{=} (F_1(s), \cdots, F_n(s))$, where $F_i \in \vec{DF}$, and $n = \|\vec{DF}\|$.

From the above two definitions it can be seen that $\forall a \in ACTION.Poss(a, s) \leftarrow_{Poss(a)} \vec{DF}$, thus pre-condition axiom (formula 2-2) can be re-written as:

$Poss(A(\vec{x}), s) \equiv \Pi_A(\vec{x}, s) \equiv \Gamma_A(\vec{x}, \vec{DF}[s])$.                    (3-4)

Informally, $\vec{PFB}$ is the maximum set of fluents which comprises all fluents that might affect the execution of some action and/or test actions, and further the system's evolution; while $\vec{DF}$ is the minimum subset of $\vec{PFB}$ that are necessary and sufficient to derive all execution possibilities at that time by querying and/or logic computing.

For a given situation calculus system, the selection of $\vec{DF}$ (and subsequently $\vec{NF}$) may have multiple choices. For example, $\vec{PFB}$ is $\{F_1, F_2, G_1, G_2, G_3\}$ and $\{F_1 \leftarrow_{F1} \{G_1\}, F_2 \leftarrow_{F2} \{\}$, $G_1 \leftarrow_{G1} \{F_1\}, G_2 \leftarrow_{G2} \{F_2\}, G_3 \leftarrow_{G3} \{F_2\}\}$. Then $\vec{DF}$ might be $\{F_1, F_2\}$ or $\{G_1, F_2\}$.

The relation between *FLUENT*, $\vec{PFB}$, $\vec{DF}$ and $\vec{NF}$ can be summarized as figure 3-1.

---

[11] Here we omit the argument of sort situation in *Poss*, *F* and *G*.

$DF_1$ and $DF_2$ are used to denote multiple choices of $\overrightarrow{DF}$.

We have a straightforward proposition:

**Proposition 3.2**      Atomic fluents in $\overrightarrow{PFB}$ are always decisive fluents.          ∎

Note that the inverse proposition does not hold. Consider the above example, $F_2$ is an atomic- and a decisive-fluent; while either $F_1$ or $G_1$ might be decisive, but neither of them is atomic since their derivation relation forms a loop.

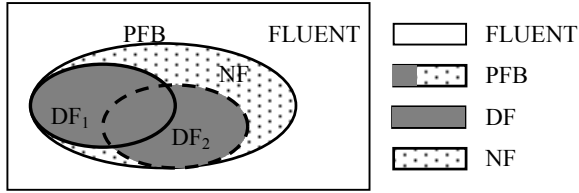

Fig.3-1      Classification of fluents

The acquisition of $\overrightarrow{DF}$ is a classical graph theory problem. Here we omit formal algorithm and just give informal description:

1)    List every action's precondition and every test action. This can be done since a situation calculus system can only have finitely many actions. Mark all pre-conditions and test actions (regarded as fluents) as *standing*.

2)    Select one *standing* fluent $f$, list all of the fluents in $f$'s every seed set. If $f$ is an atomic fluent, re-mark $f$ as *decisive*; otherwise re-mark $f$ as *non-decisive*. For all fluents in $f$'s seed sets, mark those that appear for the first time as *standing*, and leave others (those that have already been marked as *decisive*, *standing* or *non-decisive*) alone. Work on this step recursively until no *standing* fluent is left.

3)    For those *non-decisive* fluents that cannot be derived (directly or transitively) from one or more *decisive* ones, choose a *non-decisive* one and re-mark it as *decisive*. Work on this step recursively until every *non-decisive* fluent can be derived (directly or transitively) from *decisive* ones.

Then $\overrightarrow{DF}$ is composed of all fluents that marked as *decisive*, and $\overrightarrow{NF}$ is composed of all fluents marked as *non-decisive*.

Many AI and/or graph theory tricks should be employed in step 2) and 3): such as depth-first searching, width-first searching, weighted searching, domain-independent heuristic policies (such as selecting at first those seed sets that have the maximum number of seed fluents, or those that have the maximum number of atomic fluents, etc), and/or domain-dependent heuristic policies (such as selecting at first those fluents that encapsulate jumbo data storage or raw physical sensors, etc.). Interested readers can derive relevant algorithm by themselves.

## 3.4    Decisive Fluent and Situation

Suppose in a given situation calculus system: 1), $ACTION = \{a_1(\overrightarrow{x_1}), \cdots, a_m(\overrightarrow{x_m})\}$

$(m{\geq}1)$; 2) $FLUENT = \{F_1(\vec{y_1}),\cdots,F_p(\vec{y_p})\}$  $(p{\geq}1)$; 3) $TESTACTION = \{\varphi_1(\vec{z_1}),\cdots,\varphi_q(\vec{z_q})\}$  $(q{\geq}1)$;

4), $\vec{DF} = \{df_1,\cdots,df_n\}$  $(n{\geq}1)$ is one of the decisive fluent sets.

**Th. 3.1    Pre-condition and Successor State for Deterministic Complex Action**

If any $df_i$'s successor state wrt any action in ACTION is closed wrt $\vec{DF}$, for any deterministic complex action $C(a_1,...,a_m,\varphi_1,...,\varphi_q)$ whose primitive actions are from ACTION and whose test actions are from TESTACTION, there exist formulae $\Gamma$ and $\Psi$ such that:

$$Poss(C(a_1,\cdots,a_m,\varphi_1,\cdots,\varphi_q),s) \equiv$$
$$\Gamma_{C(a_1,\cdots,a_m,\varphi_1,\cdots,\varphi_q)}(a_1,\cdots,a_m,\varphi_1,\cdots,\varphi_q,\bigcup_{j=1..m}\vec{x_j},\bigcup_{k=1..q}\vec{z_k},\bigcup_{DF}\vec{y_{DFi}},\vec{DF}[s]) \text{, and} \qquad (3\text{-}5)$$

$$\forall F \in \vec{PFB}.F(\vec{y},do(C(a_1,\cdots,a_m,\varphi_1,\cdots,\varphi_q),s)) =$$
$$\Psi_{F,C(a_1,\cdots,a_m,\varphi_1,\cdots,\varphi_q)}(a_1,\cdots,a_m,\varphi_1,\cdots,\varphi_q,\vec{y},\bigcup_{j=1..m}\vec{x_j},\bigcup_{k=1..q}\vec{z_k},\bigcup_{DF}\vec{y_{DFi}},\vec{DF}[s]) \text{,} \qquad (3\text{-}6)$$

where $\Gamma$ and $\Psi$ are uniform in $s$, and free variables are among $a_1,...,a_m$, $\varphi_1,...,\varphi_q$, $\vec{x_j}$, $\vec{z_k}$, $\vec{y_i}$ and $s$.                                                                                        ∎

**Proof:**    For space reasons, the proof is listed in Appendix I.                                   ∎

Th.3.1 tells a similar conclusion to that of primitive actions: the pre-conditions and successor states can be determined in the current situation, i.e. they satisfy *Markov* property.

Th.3.2 reveals further relationship underlying between decisive fluents and situations.

**Th. 3.2**    Assuming that every decisive fluent's successor state wrt any action in ACTION is closed wrt $\vec{DF}$, situation $s_1$ and $s_2$ are evolutionally equivalent if they agree on every decisive fluent.                                                                                                                          ∎

**Analysis**:

From Th.3.1, for any deterministic complex action $\delta$, successor state of each decisive fluent $df$ can be computed by a formula $\Psi_{df,\delta}$. Then for any action $a$, its precondition $Poss(a,do(\delta,s))$ can be computable from $\vec{DF}[do(\delta,s)] = (\Psi_{df1,\delta},\cdots,\Psi_{dfn,\delta})$. All to verify is $Poss(a,do(\delta,s_1)) \equiv Poss(a,do(\delta,s_2))$ on condition that $s_1$ and $s_2$ agree on every decisive fluent.

Roughly speaking it can be done inductively with reference to the proof of Th.3.1. Detailed proof is omitted for space reasons.                                                                                ∎

## 3.5    Discussion

In this section we have explored the inherent, underlying relationship between situations and fluents, and then classified fluents according to their contribution in system's evolution.

### 3.5.1    Comparison with Current Formalisms

Our approach is built on standard (Reiter's) situation calculus formalism, and inherited much from it. Reiter's situation-action-effect formalism is the underlying reasoning mechanism supporting our decisive-fluent-set-based approach. Clearly decisive fluent set is not situation, but from evolution point of view, decisive fluent set is sufficient to determine the system future, which is usually the role of situations.

Our approach differs from fluent- or state-based approach, although they seem very alike. In fluent calculus [13], a state is composed of all relevant fluents. In state update axiom, an action's effect is represented by combination of every fluent that is possible to have positive or negative transition. We believe it is coarse to treat all fluents equally. As long as evolution is considered, the decisive fluent set, a subset of fluents, is sufficient to serve the purpose.

### 3.5.2    Merit and Shortage

Most notably, our approach utilizes incomplete knowledge, only fluents in precondition fluent base, to compute situation evolution. Moreover we exploit the underlying derivation relationship among fluents to minimize the number of fluents to form the decisive fluent set. Obviously computing and maintaining the decisive fluent set during evolution procedure is less complex than doing so on the whole FLUENT.

Another difference is that our approach inherently supports fluents with value domain of arbitrary data type, while traditional ones (such as Lin [10][11], Thielscher [13]) work well with just those fluents of Boolean type. In commercial development, functional fluents can be directly used to encapsulate[12] information sources of arbitrary complex data type, for example jumbo database tables or complex vector data from other network services. This is of much value since situation calculus will be easier to be translated into commercial development practice without extra bridging effort.

On the other hand, the decisive fluent set implies that it contains finitely many instances of fluents only, and that the complete knowledge of fluent-precondition is required. This might be a theoretical restriction compared to standard situation calculus. (So we call it a situation calculus-based) But in commercial development of network services, according to principles of software engineering, in requirement analysis phase every possible fluent, action and test-action should be listed, documented and agreed by developer and customer. Thus fluent-action relationship will have been explicitly axiomatized by end of designing phase. By the way, in fact even in some standard approaches such as fluent calculus [13], all relevant fluents, actions and the relationship between them should be enumerated before the reasoning computing becomes possible. Our approach just imposes similar restrictions.

Finally the decisive-fluent-set-based reasoning on evolution works in a similar way to that of the widely-used, traditional FSM (Finite State Machine) model. For example, decisive fluents, as well as states in FSM, can often be associated with persistent objects in physical world, for example raw data storage, physical sensors, etc. It is easy for engineers to understand, and to utilize rich FSM-based legacy methodologies and tools.

---

[12] The '*encapsulate*' refers that in Object-Oriented methodology, which is widely accepted in nowadays commercial software engineering.

# 4    Validity Theory

As mentioned earlier, autonomous computing, such PMP, is necessary for ubiquitous information services, for example PMP[13] policy. Then two basic problems arise: how to determine the correctness of autonomous computing? How to ensure the correctness of autonomous computing? The standard situation calculus does not have relevant representing and reasoning mechanism.

The solution lies in introduction of validity theory, basically formalized version of PMP policy. First we introduce *situation validity* to represent common knowledge based validity requirements, for example in $ACS^{14}$ service no pair of events in schedule could have overlapped time duration. Then we formalize validity-ensured policy to comprise all available execution choices in an organic way.

Heuristically there are at least four execution choices: *raw execution*, which will do just as requested, no less and no more; *supplement policy*, which will do some extra work in addition to the raw execution; *substitution policy*, which will satisfy only part of the raw execution; and *refusal policy*, which will decline the request at all.

A paralleling example to the above execution choices might be phone call answering. When a phone rings (the request comes): the *raw execution* is that the callee picks up the phone and answer it (real-time full-duplex audio); if the callee has moved to another place and the switch is smart enough to forward the phone to the new place, this is *complement policy* (real-time full-duplex audio with automatic re-direction); if the callee is not available and the caller is asked to leave a message on the phone recorder, this is *substitution policy* (asynchronous simplex audio); if nothing will be done if the callee does not answer the phone, this is *refusal policy* (no audio communication).

The choices are attempted in priority so as to try the best to serve the raw request while retaining the validity requirements. This can be done autonomously in our calculus framework.

## 4.1    Situation Validity

**Def.5.1    Situation Validity**    A *Situation Validity Predict VA* is a relational fluent:

$$VA(\vec{y}, s) : Object^n \times Situation \rightarrow Boolean .$$

Say a situation is *valid* if and only if the predict **VA**(s) holds.                    ∎

Situation validity can be used to describe those validity requirements of *Markov* property, i.e. those depending only on the current situation, not on others. For example, in ACS service, it is not permitted to arrange events with overlapped time duration, then the *VA* can be written as:

$$VA(s) \overset{def}{\equiv} \forall ev_1, ev_2 \in agenda\_list(s). \neg time\_overlap(ev_1, ev_2) .$$

Contrarily, if the validity requirements involve multiple situations, they cannot be translated into situation validity predicates. For example, it is required that the salary of an employee *x* can not decrease during his career [17]:

---

[13]  See page 3 for PMP policy.
[14]  See page 2 for ACS service.

$$(\forall x, s_1, s_2).(s_1 < s_2 \supset salary(x, s_1) \leq salary(x, s_2)).$$

## 4.2    Action Validity

**Def.5.2    Action Validity**    Say *a deterministic complex action $\delta$ is valid wrt situation s* if and only if $do(\delta, s)$ is valid, i.e. $VA(\vec{y}, do(\delta, s)) = True$ .    ∎

The underlying meaning is straightforward: an action is valid if and only if its execution will not produce a situation that violates validity requirements. Furthermore, if action validity can be predicted prior to actual execution, the system may have opportunities to take proper measures.

We have:

**Th.4.1    Predictability of Action Validity**    For any deterministic complex action, its validity can be predicted in the current situation.    ∎

**Proof:**

Since *VA* is essentially a Boolean fluent, from Th.3.1 there exists a formula $\Psi_{VA}$ such that:

$$(\forall \delta).VA(\vec{y}, do(\delta, s)) \equiv \Psi_{VA}(\delta, \vec{y}, \bigcup_{DF} \vec{y_k}, \overrightarrow{DF}[s]),\qquad(4\text{-}1)$$

where $\delta$ is an arbitrary deterministic complex action, and $\Psi_{VA}$ is a formula uniform in *s*, i.e. computable in current situation.    ∎

## 4.3    Validity-Assured PMP Policy

To formalize PMP[15] policy, we formalize four execution choices and a Validity-Assured PMP policy comprised of the choices. The basic purpose is: (1) to prevent actions from violating validity requirements; (2) to try the best to implement the raw request; (3) to do so in a highly autonomous way.

We adopt the heuristic execution choices mentioned earlier: *raw execution*, *supplement policy*, *substitution policy* and *refusal policy*. Apparently the raw execution can be formalized as a deterministic complex action, say $\delta$.

### 4.3.1    Refusal Policy

The refusal policy is straightforward: if an action is invalid, it is refused right away:

$$Valid\_\mathrm{Rej}(\delta) \stackrel{def}{\equiv} if \ VA(do(\delta, s)) \ then \ \delta \ else \ \mathrm{Reject}(\delta) \ endif,\qquad(4\text{-}2)$$

where *Reject($\delta$)* stands for the refusal measure whose implementation varies from one application to another.

### 4.3.2    Supplement policy

The underlying idea of supplement policy is: although an action is in-valid, it is still executable if some extra adjustments are performed:

---

[15]  See page 3 for PMP policy.

$$Valid\_Sup(\delta) \overset{def}{\equiv} if \ VA(do(\delta),s) \ then \ \delta$$
$$else \ if \ VA(do([\delta;Supplement(\delta)],s)) \ then \ [\delta;Supplement(\delta)]$$
$$else \ \text{Reject}(\delta) \ endif$$
$$endif$$

(4-3)

where $Supplement(\delta)$ stands for the supplementary measure $\delta$.

### 4.3.3    Substitution Policy

The underlying idea of substitution policy is: although an action is in-valid, the raw execution intention might be *partially* implemented with a little work left for later processing:

$$Valid\_\text{Alt}(\delta) \overset{def}{\equiv} if \ VA(do(\delta),s) \ then \ \delta$$
$$else \ if \ VA(do(Altern(\delta))) \ then \ Altern(\delta)$$
$$else \ \text{Reject}(\delta) \ endif$$
$$endif$$

(4-4)

where $Altern(a)$ stands for the action to substitute $\delta$.

### 4.3.4    Validity-Assured PMP Policy

The above policies can be combined to build a Validity-Assured PMP policy in which raw execution, supplement policy and substitution policy are attempted in turn before complete refusal in the end:

$$Valid\_\text{Service}(\delta) \overset{def}{\equiv}$$
$$if \ VA(do(\delta),s) \ then \ \delta$$
$$else \ if \ VA(do([\delta;Supplemeent(\delta)],s)) \ then \ [\delta;Supplement(\delta)]$$
$$else \ if \ VA(do(Altern(\delta),s)) \ then \ Altern(\delta)$$
$$else \ if \ VA(do([Altern(\delta);Supplement(\delta)],s)) \ then \ [Altern(\delta);Supplement(\delta)]$$
$$else \ \text{Reject}(\delta) \ endif \quad endif \quad endif \quad endif$$

.(4-5)

## 4.4    Discussion

Informally the four choices vary in how much the raw request is accepted, it might be: completely accepted by raw execution; or completely accepted extra adjustment by complement policy; or partially accepted by substitution policy; or completely rejected by refusal policy. In practical development, more subtle choices can be developed and more complex Validity-Assured PMP policies can be built. Most of all, all these can be computed automatically in our calculus framework.

# 5 Example--Active Calendar Service

In this section we will demonstrate how to model the $ACS^{16}$ service, a typical ubiquitous information service, with our calculus.

## 5.1 Requirement Axiomzation

### 5.1.1 Objects

OBJECT comprises of two types: *event* and *eventset*. The latter is the set of the former.

According to [18] and as depicted in figure 1-2, an event can be summarized as a tuple: (*ID*, *start_time*, *end_time*, *priority*, *description*). Two events are considered *collided* if they have overlapped time durations:

$$collide(ev_1, ev_2) \equiv ev_1 \neq ev_2 \wedge time\_overlap(ev_1, ev_2).$$

Obviously, *time_overlap*() can be defined from the starting time and ending time of $ev_1$ and $ev_2$, and is omitted here. In fact *collide*() could have other application-specific meanings, for example no meeting should be assigned in Christmas holiday unless under extreme emergency.

Multiple events can aggregate into an event set.

### 5.1.2 Fluents

ACS has two lists to store a user's agenda events, denoted as the solid-lined rectangle and the dashed-lined one in figure 1-2, respectively: an *agenda_list* to store confirmed events, which have no collision and is ready to alert the user at appropriate time; a *tempagn_list* to store temporary events each of which collides with at least one event in *agenda_list* but has higher priority. It is the events in *tempagn_list* that might require the user to make ultimate decisions (if they will not be transferred into *agenda_list* nor deleted from ACS completely during subsequent processing).

The *agenda_list* and *tempagn_list* in situation *s* can be acquired respectively through fluents *agenda_list*(*s*) and *tempagn_list*(*s*), both with arity *SITUATION→ eventset*.

### 5.1.3 Primitive Actions, Pre-conditions and Successor States

There are four primitive actions: *add_to_agenda_list*(*ev*) to add an event into *agenda_list*, *add_to_tempagn_list*(*ev*) to add an event into *tempagn_list*; *del_fr_agenda_list*(*ev*) to delete an event from *agenda_list*, and *del_fr_tempagn_list*(*ev*) to delete an event from *tempagn_list*.

To avoid adding duplicate agendas or deleting non-existent ones, we have pre-conditions:

$$Poss(add\_to\_agenda\_list(ev), s) \equiv ev \notin agenda\_list(s) \wedge ev \notin tempagn\_list(s),$$

$$Poss(del\_fr\_agenda\_list(ev), s) \equiv ev \in agenda\_list(s),$$

$$Poss(add\_to\_tempagn\_list(ev), s) \equiv ev \notin agenda\_list(s) \wedge ev \notin tempagn\_list(s),$$

---

[16] See page 2 for ACS service.

$Poss(del\_fr\_tempagn\_list(ev),s) \equiv ev \in tempagn\_list(s)$ .

We have the following successor state axioms:

$agenda\_list(do(add\_to\_agenda\_list(ev),s)) \equiv agenda\_list(s) \cup \{ev\}$ ,

$agenda\_list(do(del\_fr\_agenda\_list(ev),s)) \equiv agenda\_list(s) - \{ev\}$ ,

$agenda\_list(do(add\_to\_tempagn\_list(ev),s)) \equiv tempagn\_list(s) \cup \{ev\}$ ,

$agenda\_list(do(del\_fr\_tempagn\_list(ev),s)) \equiv tempagn\_list(s) - \{ev\}$ .

It is easy to verify that the above successor states are closed wrt to {*agenda_list*(*s*), *tempagn_list*(*s*)} (in fact they are incremental), and that they satisfy formula 3-6 in Th.3.1.

### 5.1.4    Situation Validity

As mentioned earlier, at any time no collision is allowed in *agenda_list*, while each event in *tempagn_list* should conflict with at least one in *agenda_list* but has higher priority. With an auxiliary situation-independent function *collide_with_list* and fluents *VA1* and *VA2*, situation validity *VA* can be defined as:

$collide\_with\_list(ev,list) \overset{def}{=} \exists b \in list.ev \neq b \wedge collide(ev,b)$ ,

$VA1(s) \overset{def}{\equiv} (\forall ev_1, ev_2 \in agenda\_list(s)).(ev_1 \neq ev_2) \supset \neg collide(ev_1, ev_2)$ ,

$VA2(s) \overset{def}{\equiv} \forall ev \in tempagn\_list(s).[collide\_with\_list(ev, agenda\_list(s) \wedge$
$\quad (\forall b \in agenda\_list(s).[collide(ev,b) \supset ev.priority > b.priority]) \wedge$
$\quad (\forall c \in tempagn\_list(s).$
$\qquad [ev \neq c \wedge collide(ev,c)) \supset ev.priority \geq c.priority]]$ ,

$VA(s) \overset{def}{\equiv} VA1(s) \wedge VA2(s)$ .

### 5.1.5    Validity-Assured PMP Policy

ACS exploits Validity-Assured PMP policy as formula 4-5. Every service has its specific execution choices of raw service, refusal, supplement and substitution policies.

**(1) To add a new agenda**

The raw execution is *add_to_agenda_list*(*ev*).

*Reject*(*add_to_agenda_list*(*ev*)) will not affect either list, its implementation is application-specific and is beyond situation calculus. For example, it might be to send negative acknowledgement to the requestor.

*Supplement*() and some auxiliary functions are defined in figure 5-1. Intuitively, since it is the non-empty *tempagn_list* that might require the user's manual intervention, *Supplement*() tries to minimize *tempagn_list* either by transferring temporary events into *agenda_list* or by dropping them out of ACS at all. Since the transferring or dropping of an event might probably cause other events become possible to transfer or to drop, the procedure will be done recursively.

For *Altern*(*add_to_agenda_list*(*ev*)), if an event cannot be put into *agenda_list* at this time, it will be temporarily stored in *tempagn_list*. Thus

$$Altern(add\_to\_agenda\_list(ev)) \overset{def}{\equiv} add\_to\_tempagn\_list(ev).$$

**(2) To delete an agenda**

The raw execution is [*del_fr_agenda_list*(*ev*)>>*del_fr_tempagn_list*(*ev*)].

*Reject*([*del_fr_agenda_list*(*ev*)>>*del_fr_tempagn_list*(*ev*)]) does not affect either list, its implementation is application-specific and is out of situation calculus. For example, it might send negative acknowledgement to the requestor.

The supplement policy is described as in figure 5-1.

No substitution policy is required.

### 5.1.6    Decisive Fluents

*agenda_list*(*s*) and *tempagn_list*(*s*) are atomic fluents, and are used to derive pre-conditions and test actions in *VA1*, *VA2* and *Supplement*. Thus they are decisive fluents. No other decisive fluent can be found. Thus $\overrightarrow{DF}$ ={*agenda_list*, *tempagn_list*}.

$Supplement() \overset{def}{\equiv}$

$while \ \neg VA2()$

$do \quad [$

    $while \ \exists ev \in tempagn\_list().\neg collide\_with\_list(ev, agenda\_list()) \wedge$

           $\neg collide\_with\_list(ag, tempagn\_list())$

    $do \quad [del\_fr\_tempagn\_list(ev); add\_to\_agenda\_list(ev)]$

    $endwhile$

    $>>$

    $while \ \exists ev \in tempagn\_list().\neg collide\_with\_list(ev, agenda\_list()) \wedge$

           $collide\_with\_list(ev, tempagn\_list())$

    $do \quad if \ \ highest\_prio(ev, collide\_sublist(ev, tempagn\_list()))$

        $then$

        $[ \ while \ \ collide\_sublist(ev, tempagn\_list()) \neq \{\}$

          $do \quad \rho\pi(b \in collide\_sublist(ev, tempagn\_list()).$

               $del\_fr\_tempagn\_list(b)$

          $endwhile \ ;$

          $del\_fr\_tempagn\_list(ev) \ ; \ add\_to\_agenda\_list(ev)$

        $]$

        $endif$

    $endwhile$

    $>>$

    $while \ \exists ev \in tempagn\_list().collide\_with\_list(ev, agenda\_list()) \wedge$

           $collide\_with\_list(ev, tempagn\_list())$

    $do \quad if \ \ highest\_prio(ev, collide\_sublist(ev, tempagn\_list()))$

        $then$

        $while \ \ collide\_sublist(ev, tempagn\_list()) \neq \{\}$

        $do \quad \rho\pi(b \in collide\_sublist(ev, tempagn\_list()).$

            $del\_fr\_tempagn\_list(b)$

        $endwhile$

        $endif$

    $endwhile$

    $]$

$endwhile$

        Fig.5-1      supplement policy

## 5.2    Scenario

The exemplar scenario implemented here is $t_0$-$t_1$-$t_2$-$t_3$ in figure 1-2: first to add two new agendas *ev003* and *ev004* successively, then to delete *ev001*. The Validity-Assured PMP policy as formula 4-5 will be exploited at each step.

### 5.2.1    Initial Situation

At $t_0$, $\overrightarrow{DF}[s_0]$ = (agenda_list($S_0$), temp_agnlist($S_0$)) = ({*ev001, ev002*}, {}), where *ev001* and *ev002* are depicted in figure 1-2.

### 5.2.2    Adding New Events

Now come successively two requests to add *ev003* and *ev004*. Obviously *ev003* conflicts with *ev002*, and *ev004* conflicts with *ev001*.

To add *ev003*, consider the pre-condition:

$$Poss(add\_to\_agenda\_list(ev003), S_0)$$
$$= ev003 \notin agenda\_list(S_0) \wedge ev003 \notin tempagn\_list(S_0) = True$$.

Then consider each *if*- branch of the validity-assured PMP policy as formula 4-5. Raw execution and supplement policy will be aborted since

$$VA(do(add\_to\_agenda\_list(ev003), S_0))$$
$$= VA((agenda\_list(S_0) \bigcup \{ev003\},\ tempagn\_list(S_0))) = False$$ and

$$VA(do([add\_to\_agenda\_list(ev003); Supplement\,()], S_0))$$
$$= VA((agenda\_list(S_0) \bigcup \{ev003\}, tempagn\_list(S_0))) = False$$.

For substitution policy, since

$$VA(do(add\_to\_tempagn\_list(ev003), S_0))$$
$$= VA((agenda\_list(S_0),\ tempagn\_list(S_0) \bigcup \{ev003\})) = True$$,

it is executable, and the result is:

$$\overrightarrow{DF}[s_1] = (agenda\_list(s_0), temp\_agnlist(s_0) \bigcup \{ev003\})$$
$$= (\{ev001, ev002\}, \{ev003\})$$.

Similarly, *ev004* can be added through substitution policy, and the result is:

$$\overrightarrow{DF}[s_2] = (agenda\_list(s_1), temp\_agnlist(s_1) \bigcup \{ev004\})$$
$$= (\{ev001, ev002\}, \{ev003, ev004\})$$.

### 5.2.3    Deleting Existing Agendas

Now comes a request to delete *ev001*. Consider the pre-condition:

$$Poss([del\_fr\_agenda\_list(ev001) >> del\_fr\_tempagn\_list(ev001)], s_1)$$
$$= ev001 \in agenda\_list(s_1) \lor ev001 \in tempagn\_list(s_1) = True$$ .

For raw execution will be aborted since

$$VA(do([del\_fr\_agenda\_list(ev001) >> del\_fr\_tempagn\_list(ev001)], s_2))$$
$$= VA((agenda\_list(s_2) - \{ev001\}, tempagn\_list(s_2))) = False$$ .

For supplement policy, since

$$VA(do([[del\_fr\_agenda\_list(ev001) >> del\_fr\_tempagn\_list(ev001)]; Supplement()], s_2))$$
$$= VA((agenda\_list(s_2) - \{ev001\} \cup \{ev004\}, tempagn\_list(s_2) - \{ev004\})) = True$$ ,

it is executable. Then the result is:

$$\overrightarrow{DF}[s_3] = (\{ev004, ev002\}, \{ev003\}).$$

It can seen that when *ev001* is deleted, *ev004* will be automatically transferred into *agenda_list* by supplement policy, and free from the user's intervention.

# 6   Summary

With honorable, proven success in modeling dynamical world in artificial intelligence research as well as in some practical problems, situation calculus is chosen to model ubiquitous information services for commercial development purpose. Based on Reiter's standard formalism, we made some augmentation: 1) exploration on fluents' role in evolution; 2) introduction of validity theory to support autonomous, validity-ensured, multiple-choices-enabled services.

One essential role of situation is to mark starting points for future evolution. This can also be done by the decisive fluent set, a subset of FLUENT. All fluents that might affect some action's precondition or some test actions are collected into precondition fluent base, whose seed fluents are further collected as decisive fluent sets. A decisive fluent set is sufficient and necessary to determine any evolution potentials. Noticably a decisive fluent set is just incomplete knowledge in a situation. Besides since fluents can be of any data type and the decisive fluent set works similar way to FSM, our approach is easy to translate into commercial development.

To allow autonomous computing, ubiquitous services should exploit alternative choices among full acceptance, partial acceptance and full rejection to serve the requests. Validity theory is introduced to serve the formalization purpose: Validity fluents are used to axiomatize application-specific validity requirements. Validity-ensured policy comprises the multiple alternative execution choices into an organic execution unit. With validity-ensured policy, computers can autonomously select an execution choice to best serve the raw service request while retaining validity requirements. Notice that validity theory allows more than simply acceptance or rejection.

Our work makes a fine starting point to model ubiquitous applications, but is yet far from perfect. Since the formal calculus itself is hard for practical engineers to understand, is hard to be translated into commercial development documents, and is not compatible with mainstream commercial development platforms, further work has to be done like [19][20].

# 7   Reference

[1]. Mark Weiser. (1996) Some Computer Science Problems in Ubiquitous Computing. Communications of the ACM, 36, 74-84.

[2]. Weikai XIE, Guangyou XU, Yuanchun SHI. (2002) Ubiquitous Computing--Computing Paradigm of The Next Generation (in Chinese). China Computer World, 8, B1-B3.

[3]. Imielinski T, Goel S. (2000) DataSpace: Querying and Monitoring Deeply Networked Collections In Physical Space. IEEE Personal Communications, 7, 4-9.

[4] Wenyu DONG, Shengbin CUI, Xin GAO, Xiaohong DONG. (2002) LWIS--A Location-dependent Wireless Information Service Model (in Chinese). (China) Computer Engineering and Application, 38, 153-155.

[5]  Claudio S. Pinhanez, Aaron F. Bobick. (2003) Interval scripts: A Programming Paradigm for Interactive Environments and Agents. Personal and Ubiquitous Computing, 7, 1-21.

[6]  J. McCarthy (1968). Situations, Actions and Causal Laws. In M. Minsky (eds), Semantic Information Processing. MIT Press, Cambridge, Mass.

[7]  Fiora Pirri, Ray Reiter. (1999) Some Contributions To The Metatheory of the Situation Calculus. Journal Of ACM, 46, 325-364.

[8]  Y. Lesperance, H. J. Levesque, R. Reiter. (1997) A Situation Calculus Approach To Modeling And Programming Agents. In A. Rao and M. Wooldridge (eds), Foundations And Theories Of Rational Agency. Kluwer, New York.

[9]  Yves Lesperance, Hector Levesque, Fangzhen Lin, Daniel Marcu, Raymond Reiter, and Richard Scherl. (1994) A Logical Approach To High-level Robot Programming--A Progress Report. In Benjamin Kuipers (eds.), Control of the Physical World by Intelligent Systems: Papers from the 1994 AAAI Fall Symposium. New Orleans, LA, Nov. 1994. 79-85. AAAI Press, Menlo Park, CA, USA.

[10] F. Lin and R. Reiter (1994). How to Progress A Database (and Why) I: Formal Foundations. In Jon Doyle, Erik Sandwall and Pietro Torasso (eds.), Proceedings of KR'94, Fourth International Conference on Principles of Knowledge Representation and Reasoning. Bonn, Germany, 1994. 425-436. Morgan Kaufmann, Los Altos.

[11] F. Lin and R. Reiter. (1995) How to progress a database II: The Strips Connection. In Proceedings on Fourteenth International Joint Conference on Artificial Intelligence (IJCAI-95). August, 1995. Montreal, Canada. 2001-2007. Morgan Kaufmann Publishers, San Francisco, USA.

[12] Hector J. Levesque, Raymond Reiter, Yveslesperance, Fangzhen Lin, and Richard B. Scherl. (1997) GOLOG: A Logic Programming Language For Dynamic Domains, Journal Of Logic Programming, 31, 59-84.

[13] Michael Thielscher. (1999) From Situation Calculus To Fluent Calculus: State Update Axioms As A Solution To The Inferential Frame Problem. Artificial Intelligence, 111, 277-299.

[14] Michael Thielscher. (1998) Introduction To The Fluent Calculus. Linkoping Electronic Articles in Computer and Information Science. 3(014). http://www.ep.liu.se/ea/cis/1998/014/.

[15] John McCarthy (2002) Actions And Other Events In Situation Calculus. 8[th] International Conference on Principles of Knowledge Representation and Reasoning (KR2002), Toulouse, France, 22-25 April. Morgan Kaufmann Publishers, San Francisco, USA.

[16] John McCarthy and Patrick J. Hayes (1969) Some Philosophical Problems From The

Standpoint Of Artificial Intelligence. In Bernard Meltzer and Donald Michie (eds), Machine Intelligence 4. pp. 463-502. Edinburgh University Press, Edinburgh, UK.

[17] Ray Reiter (1993) Proving Properties of States in the Situation Calculus. *Artificial Intelligence*, 64,337-351.

[18] Roland H. Alden, et al. (1998) vCalendar--The Electronic Calendaring and Scheduling Exchange Format, v1.0. http://www.imc.org/pdi/.

[19] Leopoldo Bertossi, Marcelo Arenas, Cristian Ferretti. (1998) SCDBR: An Automated Reasoner for Specifications of Database Updates. Journal of Intelligent Information Systems, 10, 253--280.

[20] Wenyu DONG, Donghong SUN, Ke XU, Xuedong LI. Modeling of Autonomous Network Information Service (in Chinese, submitted).

# Appendix I    Proof of Theorem 3.1

In this appendix we are going to prove Theorem 3.1.

Suppose in a given situation calculus system: 1), $ACTION = \{a_1(\vec{x_1}), \cdots, a_m(\vec{x_m})\}$

$(m \geq 1)$; 2) $FLUENT = \{F_1(\vec{y_1}), \cdots, F_p(\vec{y_p})\}$  $(p \geq 1)$; 3) $TESTACTION = \{\varphi_1(\vec{z_1}), \cdots, \varphi_q(\vec{z_q})\}$  $(q \geq 1)$;

4), $\overrightarrow{DF} = \{df_1, \cdots, df_n\}$  $(n = \left\|\overrightarrow{DF}\right\|)$ is one of the decisive fluent sets.

## Th. 3.1  Pre-condition and Successor State for Deterministic Complex Action

If any $df_i$'s successor state wrt any action in ACTION is close wrt $\overrightarrow{DF}$, for any deterministic complex action $C(a_1, \ldots, a_m, \varphi_1, \ldots, \varphi_q)$ whose primitive actions are from ACTION and whose test

actions are from TESTACTION, there exist formulae $\Gamma$ and $\Psi$ such that:

$$Poss(C(a_1, \cdots, a_m, \varphi_1, \cdots, \varphi_q), s) \equiv$$
$$\Gamma_{C(a_1, \cdots, a_m, \varphi_1, \cdots, \varphi_q)}(a_1, \cdots, a_m, \varphi_1, \cdots, \varphi_q, \bigcup_{j=1..m}\vec{x_j}, \bigcup_{k=1..q}\vec{z_k}, \bigcup_{DF}\overrightarrow{y_{DFi}}, \overrightarrow{DF}[s]) \text{, and} \quad (3\text{-}5)$$

$$\forall F \in \overrightarrow{PFB}.F(\vec{y}, do(C(a_1, \cdots, a_m, \varphi_1, \cdots, \varphi_q), s)) =$$
$$\Psi_{F, C(a_1, \cdots, a_m, \varphi_1, \cdots, \varphi_q)}(a_1, \cdots, a_m, \varphi_1, \cdots, \varphi_q, \vec{y}, \bigcup_{j=1..m}\vec{x_j}, \bigcup_{k=1..q}\vec{z_k}, \bigcup_{DF}\overrightarrow{y_{DFi}}, \overrightarrow{DF}[s]) \text{,} \quad (3\text{-}6)$$

where $\Gamma$ and $\Psi$ are uniform in $s$, and free variables are among $a_1, \ldots, a_m, \varphi_1, \ldots, \varphi_q$, $\vec{x_j}$, $\vec{z_k}$, $\vec{y_i}$

and $s$.                                                                                               ∎

**Proof**:

## I    Preliminaries:

**I.1**   Recall that $\overrightarrow{DF}[s] \overset{def}{=} (df_1(s), \cdots, df_n(s))$.

**I.2**   From definition of decisive fluent and pre-condition fluent base, for any primitive action or

test action, there exists a formula such that:

$$\forall a(\vec{x}) \in ACTION.Poss(a, s) = \Gamma_a(\vec{x}, \overrightarrow{DF}[s]) \text{, or} \quad (A\text{-}1)$$

$$\forall \delta(\vec{x}) \in TESTACTION.\delta(\vec{x})[s] = \Gamma_\delta(\vec{x}, \overrightarrow{DF}[s]) \text{,} \quad (A\text{-}2)$$

where $\Gamma_a$ and $\Gamma_\delta$ are uniform in $s$ with free variables among $\vec{x}$ and $s$.

Since any $df$ is closed wrt $\overrightarrow{DF}$, for any decisive fluent $df$, there exists a formula $\Psi$ such that:

$$\forall df \in \overrightarrow{DF}, \forall a(\vec{x}) \in ACTION.df(\vec{y}, do(a(\vec{x}), s)) = \Psi_{df}(\vec{y}, \vec{x}, a, \overrightarrow{DF}[s]) \text{,} \quad (A\text{-}3)$$

where $\Psi_{df}$ is uniform in $s$ with free variables among $\vec{x}$, $\vec{y}$, $a$ and $s$.

**I.3**   If for any decisive fluent $dfi$ there exists a formula $\Psi_{dfi,C}$ such that:

$$df_i(\vec{y}, do(C(a_1, \cdots, a_2, \delta_1, \cdots, \delta_n), s)) = \Psi_{dfi,C}(a_1, \cdots, a_m, \delta_1, \cdots, \delta_q, \bigcup_{j=1..m}\vec{x_j}, \bigcup_{k=1..q}\vec{z_k}, \bigcup_{DF}\vec{y_i}, \overrightarrow{DF}[s])\;,$$

where $\Psi_{dfi,C}$ is uniform in $s$ with free variables among $a_1, \ldots, a_m, \delta_1, \ldots, \delta_q,$ $\vec{x_1}$, $\vec{z_k}$, $\vec{y_i}$ and $s$.

Since any non-decisive fluent, say $F$, can be derived from $\overrightarrow{DF}$, $F$'s successor state will be:

$$F(\vec{y}, s) = func(df_1(\vec{y_1}, s), \cdots, df_n(\vec{y_n}, s)) = func(\Psi_{df1,C}(\cdots), \cdots, \Psi_{dfn,C}(\cdots))$$

$$\overset{def}{=} \Psi_{F,C}(a_1, \cdots, a_m, \delta_1, \cdots, \delta_q, \bigcup_{j=1..m}\vec{x_j}, \bigcup_{k=1..q}\vec{z_k}, \bigcup_{DF}\vec{y_i}, \overrightarrow{DF}[s])\;,$$

where $\Psi_{F,C}$ is uniform in $s$ with free variables among $a_1, \ldots, a_m, \delta_1, \ldots, \delta_q,$ $\vec{x_1}$, $\vec{z_k}$, $\vec{y_i}$ and $s$.

Thus formula 3-6 is only required to verify for each decisive fluent.

**I.4**   Consider the six types of deterministic complex actions in §3.1. Informally,

For (1) primitive actions: formula A-1 and A-3 tells they conform Th.3.1.

For (2) test actions: since their truth values can be derived from formula A-2, and no fluents will be affected by their execution, they conform to Th.3.1.

For (4) deterministic choice of action arguments: recall that $[\rho\pi(\vec{x}).\delta(\vec{x})]$ stands for

$\delta(\vec{x_1}) >> \cdots >> \delta(\vec{x_n}) >> \cdots$. Informally, if each $\delta(\vec{x_i})$'s precondition and successor state

conform to Th.3.1, those of $[\rho\pi(\vec{x}).\delta(\vec{x})]$ will also do.

For (6) deterministic iterations: Informally they are combination of sequences and test actions.

Thus all that need to prove are: (3) sequences and (4) deterministic choice of actions and (6) deterministic iterations.

Since deterministic complex actions are defined recursively, we exploit inductive proof.

## II   Inductive Foundation:

**(1)**   **$[a_1; a_2]$**, where $a_1$ and $a_2$ are primitive actions with $\vec{x_i}$  ($i$=1,2) as arguments of each.

Form formula A-1 and A-3, we have:

$$\forall df_i \in \overrightarrow{DF}.df_i(\vec{y_i}, do(a_1(\vec{x_1}), s)) = \Psi_{dfi,a1}(\vec{y_i}, \vec{x_1}, \overrightarrow{DF}[s])\;.$$

The pre-condition of $[a_1; a_2]$ is:

$$Poss([a_1; a_2], s) \equiv Poss(a_1, s) \wedge Poss(a_2, do(a_1, s))$$

$$\equiv \Gamma_{a1}(\vec{x_1}, \overrightarrow{DF}[s]) \wedge \Gamma_{a2}(\vec{x_2}, df_1(\vec{y_1}, do(a_1, s)), \cdots, df_n(\vec{y_n}, do(a_1, s)))$$

$$\equiv \Gamma_{a1}(\vec{x_1}, \overrightarrow{DF}[s]) \wedge \Gamma_{a2}(\vec{x_2}, \Psi_{df1,a1}(\vec{y_1}, \vec{x_1}, \overrightarrow{DF}[s]), \cdots, \Psi_{dfn,a1}(\vec{y_n}, \vec{x_1}, \overrightarrow{DF}[s]))\;.\quad \text{(A-4)}$$

$$\overset{def}{\equiv} \Gamma_{[a1;a2]}(a_1, a_2, \vec{x_1}, \vec{x_2}, \bigcup_{DF}\vec{y_i}, \overrightarrow{DF}[s])$$

Obviously $\Gamma_{[a1;a2]}$ is uniform in $s$ with free variables among $a_1, a_2, \vec{x_1}, \vec{x_2}, \vec{y_i}$ and $s$.

For successor state for a decisive fluent $df_i$, we have:

$$df_i(do([a_1(\vec{x_1});a_2(\vec{x_2})],s))$$
$$= \Psi_{dfi,a2}(\vec{x_2}, df_1(\vec{y_1}, do(a_1(\vec{x_1}),s)),\cdots,df_n(\vec{y_n}, do(a_1(\vec{x_1}),s)))$$
$$= \Psi_{dfi,a2}(\vec{x_2}, \Psi_{df1,a1}(\vec{y_1}, \vec{x_1}, DF[s]),\cdots,\Psi_{dfn,a1}(\vec{y_1}, \vec{x_1}, DF[s])) \qquad \text{(A-5)}$$
$$\overset{def}{=} \Psi_{dfi,[a1;a2]}(a_1, a_2, \vec{x_1}, \vec{x_2}, \bigcup_{DF}\vec{y_i}, DF[s])$$

Obviously $\Psi_{dfi,[a1;a2]}$ is uniform in $s$ with free variables among $a_1, a_2, \vec{x_1}, \vec{x_2}, \vec{y_i}$ and $s$.

**(2)** **[$a_1 \gg a_2$]**, where $a_1$ and $a_2$ are primitive actions with $\vec{x_i}$ ($i$=1,2) as arguments of each.

For precondition we have:

$$Poss([a_1 \gg a_2],s) \equiv Poss(a_1,s) \lor Poss(a_2,s) \equiv \Gamma_{a1}(\vec{x_1}, \overrightarrow{DF}[s]) \lor \Gamma_{a2}(\vec{x_2}, \overrightarrow{DF}[s])$$
$$\overset{def}{\equiv} \Gamma_{[a1>a2]}(a_1, a_2, \vec{x_1}, \vec{x_2}, \overrightarrow{DF}[s]) \qquad \text{(A-6)}$$

Obviously $\Gamma_{[a1\gg a2]}$ is uniform in s with free variables among $a_1, a_2, \vec{x_1}, \vec{x_2}$ and $s$.

For a decisive fluent $df$, its successor state is:

$$df(\vec{y}, do([a_1(\vec{x_1}) \gg a_2(\vec{x_2})],s)) \equiv \begin{cases} df(\vec{y}, do(a_1(\vec{x_1}),s)) & ,if\ Poss(a_1,s) \\ df(\vec{y}, do(a_2(\vec{x_2}),s)) & ,if\ \neg Poss(a_1,s) \land Poss(a_2,s) \end{cases}.$$

Since $df(y,do(a_i,s))$ and $Poss(a_i,s)$ ($i$=1,2) are both uniformed in $s$ with free variables among $a_1, a_2, \vec{x_1}, \vec{x_2}, \vec{y}$ and $s$. It is reasonable to re-write the above formula as:

$$df(\vec{y}, do([a_1(\vec{x_1}) \gg a_2(\vec{x_2})],s)) \overset{def}{\equiv} \Psi_{df,[a1\gg a2]}(a_1, a_2, \vec{x_1}, \vec{x_2}, \vec{y}, \overrightarrow{DF}[s]), \qquad \text{(A-7)}$$

where $\Psi_{df,[a1\gg a2]}$ is uniform in s with free variables among $a_1, a_2, \vec{x_1}, \vec{x_2}$ and $s$.

## III Induction:

**(3)** **[$\delta_1;\delta_2$]**, where $\delta_1$ and $\delta_2$ are deterministic complex actions that conform to Th3.1.

Assume the pre-conditions and successor states are:

$$Poss(\delta_i(\vec{x_i}),s) \equiv \Gamma_{\delta i}(\delta_i, \vec{x_i}, \bigcup_{DF}\vec{y_k}, \overrightarrow{DF}[s]) \quad (i=1..2), \text{ and}$$

$$\forall df_j \in \overrightarrow{DF}.df_j(\vec{y_j}, do(\delta_i(\vec{x_i}),s)) = \Psi_{dfj,\delta i}(\delta_i, \vec{x_i}, \bigcup_{DF}\vec{y_k}, \overrightarrow{DF}[s]) \quad (i=1..2),$$

where $\Gamma_{\delta i}$ and $\Psi_{dfj,\delta i}$ are uniform in $s$ with free variables among $\delta_i, \vec{x_i}, \bigcup_{DF}\vec{y_k}$ and $s$.

The pre-condition of [$\delta_1;\delta_2$] is:

$Poss([\delta_1; \delta_2], s) \equiv Poss(\delta_1, s) \wedge Poss(\delta_2, do(\delta_1, s))$

$\equiv \Gamma_{\delta 1}(\delta_1, \vec{x_1}, \bigcup_{DF} \vec{y_k}, \overrightarrow{DF}[s]) \wedge \Gamma_{\delta 2}(\delta_2, \vec{x_2}, \bigcup_{DF} \vec{y_k}, df_1(\vec{y_1}, do(\delta_1(\vec{x_1}), s)), \cdots, df_n(\vec{y_n}, do(\delta_1(\vec{x_1}), s)))$

$\equiv \Gamma_{\delta 1}(\delta_1, \vec{x_1}, \bigcup_{DF} \vec{y_k}, \overrightarrow{DF}[s])$

$\qquad \wedge \Gamma_{\delta 2}(\delta_2, \vec{x_2}, \bigcup_{DF} \vec{y_k}, \Psi_{df1, \delta 1}(\delta_1, \vec{x_1}, \bigcup_{DF} \vec{y_k}, \overrightarrow{DF}[s]), \cdots, \Psi_{dfn, \delta 1}(\delta_1, \vec{x_1}, \bigcup_{DF} \vec{y_k}, \overrightarrow{DF}[s]))$

$\overset{def}{\equiv} \Gamma_{[\delta 1; \delta 2]}(\delta_1, \delta_2, \vec{x_1}, \vec{x_2}, \bigcup_{DF} \vec{y_k}, \overrightarrow{DF}[s])$

$$(A\text{-}8)$$

Obviously $\Gamma_{[\delta 1; \delta 2]}$ is uniform in $s$ with free variables among $\delta_1, \delta_2,$ $\vec{x_1}$, $\vec{x_2}$, $\bigcup_{DF} \vec{y_k}$ and $s$.

For successor state for a decisive fluent $df_i$, we have:

$df_i(\vec{y_i}, do([\delta_1(\vec{x_1}); \delta_2(\vec{x_2})], s)) = \Psi_{dfi, \delta 2}(\delta_2, \vec{x_2}, \bigcup_{DF} \vec{y_k}, \overrightarrow{DF}[do(\delta_1(\vec{x_1}), s)])$

$= \Psi_{dfi, \delta 2}(\delta_2, \vec{x_2}, \bigcup_{DF} \vec{y_k}, df_1(\delta_1, \vec{x_1}, \bigcup_{DF} \vec{y_k}, \overrightarrow{DF}[s]), \cdots, df_n(\delta_1, \vec{x_1}, \bigcup_{DF} \vec{y_k}, \overrightarrow{DF}[s]))$

$= \Psi_{dfi, \delta 2}(\delta_2, \vec{x_2}, \bigcup_{DF} \vec{y_k}, \Psi_{dfi, \delta 1}(\delta_1, \vec{x_1}, \bigcup_{DF} \vec{y_k}, \overrightarrow{DF}[s]), \cdots, \Psi_{dfi, \delta 1}(\delta_1, \vec{x_1}, \bigcup_{DF} \vec{y_k}, \overrightarrow{DF}[s]))$

$\overset{def}{=} \Psi_{dfi, [\delta 1; \delta 2]}(\delta_1, \delta_2, \vec{x_1}, \vec{x_2}, \bigcup_{DF} \vec{y_k}, \overrightarrow{DF}[s]))$

$$(A\text{-}9)$$

Obviously $\Psi_{dfi, [\delta 1; \delta 2]}$ is uniform in $s$ with free variables among $\delta_1, \delta_2,$ $\vec{x_1}$, $\vec{x_2}$, $\bigcup_{DF} \vec{y_k}$ and $s$.

**(4) $[\delta_1 >> \delta_2]$**, where $\delta_1$ and $\delta_2$ are deterministic complex actions that conform to Th.3.1.

For precondition:

$Poss([\delta_1 >> \delta_2], s) \equiv Poss(\delta_1, s) \vee Poss(\delta_2, s) \equiv \bigvee_{i=1,2} \Gamma_{\delta i}(\delta_i, \vec{x_i}, \bigcup_{DF} \vec{y_k}, \overrightarrow{DF}[s])$

$\overset{def}{\equiv} \Gamma_{[\delta 1 >> \delta 2]}(\delta_1, \delta_2, \vec{x_1}, \vec{x_2}, \bigcup_{DF} \vec{y_k}, \overrightarrow{DF}[s])$    $(A\text{-}10)$

Obviously $\Gamma_{[\delta 1 >> \delta 2]}$ is uniform in $s$ with free variables among $\delta_1, \delta_2,$ $\vec{x_1}$, $\vec{x_2}$, $\bigcup_{DF} \vec{y_k}$ and $s$.

For successor state for a decisive fluent $df$:

$df(\vec{y}, do([\delta_1(\vec{x_1}) >> \delta_2(\vec{x_2})], s)) \equiv \begin{cases} df(\vec{y}, do(\delta_1(\vec{x_1}), s)) & , if\ Poss(\delta_1, s) \\ df(\vec{y}, do(\delta_2(\vec{x_2}), s)) & , if\ \neg Poss(\delta_1, s) \wedge Poss(\delta_2, s) \end{cases}$

Since $df(y, do(\delta_i, s))$ and $Poss(\delta_i, s)$ ($i$=1,2) are both uniformed in $s$ with free variables among $\delta_1, \delta_2,$ $\vec{x_1}$, $\vec{x_2}$, $\bigcup_{DF} \vec{y_k}$ and $s$. It is reasonable to re-write the above formula as:

$df(\vec{y}, do([\delta_1(\vec{x_1}) >> \delta_2(\vec{x_2})], s)) \equiv \Psi_{df, [\delta 1 >> \delta 2]}(\delta_1, \delta_2, \vec{x_1}, \vec{x_2}, \bigcup_{DF} \vec{y_k}, \overrightarrow{DF}[s])$,    $(A\text{-}11)$

where $\Psi_{df,[\delta1>>\delta2]}$ is uniform in s with free variables among $\delta_1, \delta_2$, $\overrightarrow{x_1}$ , $\overrightarrow{x_2}$ , $\bigcup_{DF} \overrightarrow{y_k}$ and $s$.

**IV  Conclusion**:

From **I**, **II** and **III**, it can be concluded that for any deterministic complex action, the pre-condition and successor state are computable with formulae 3-5 and 3-6, which are uniform in the current situation.                                                                ■